\documentclass{article}

\usepackage{arxiv}

\usepackage[utf8]{inputenc} 
\usepackage[T1]{fontenc}    
\usepackage{hyperref}       
\usepackage{url}            
\usepackage{booktabs}       
\usepackage{amsfonts}       
\usepackage{nicefrac}       
\usepackage{microtype}      
\usepackage{lipsum}		
\usepackage{graphicx}
\usepackage{natbib}
\usepackage{doi}
\usepackage{array} 
\usepackage{booktabs} 
\usepackage{multirow}
\usepackage{graphicx}

\title{A Sentiment Analysis of Medical Text Based on Deep Learning}

\author{ \href{https://orcid.org/0000-0000-0000-0000}{\includegraphics[scale=0.06]{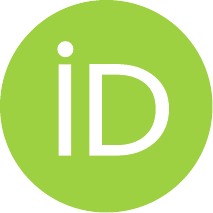}\hspace{1mm}Yinan~Chen}\\
	College of Big Data and Software\\
	Chongqing University, China\\
	\texttt{chenyinan@cqu.edu.cn} \\
}

\renewcommand{\shorttitle}{\textit{arXiv} Template}

\hypersetup{
pdftitle={A Sentiment Analysis of Medical Text Based on Deep Learning},
pdfsubject={Computer Science, Artificial intelligence},
pdfauthor={Yinan Chen},
pdfkeywords={Medical text, Sentiment Analysis, Deep learning},
}

\begin{document}
\maketitle

\begin{abstract}
	
The field of natural language processing (NLP) has made significant progress with the rapid development of deep learning technologies. One of the research directions in text sentiment analysis is sentiment analysis of medical texts, which holds great potential for application in clinical diagnosis. However, the medical field currently lacks sufficient text datasets, and the effectiveness of sentiment analysis is greatly impacted by different model design approaches, which presents challenges. Therefore, this paper focuses on the medical domain, using bidirectional encoder representations from transformers (BERT) as the basic pre-trained model and experimenting with modules such as convolutional neural network (CNN), fully connected network (FCN), and graph convolutional networks (GCN) at the output layer. Experiments and analyses were conducted on the METS-CoV dataset to explore the training performance after integrating different deep learning networks. The results indicate that CNN models outperform other networks when trained on smaller medical text datasets in combination with pre-trained models like BERT. This study highlights the significance of model selection in achieving effective sentiment analysis in the medical domain and provides a reference for future research to develop more efficient model architectures.

\end{abstract}


\section{Introduction}

Sentiment analysis (SA) is a major area of research in natural language processing that aims to identify and analyze subjective information in text data to determine the emotional sentiment expressed. This analysis can identify whether the text conveys positive, negative, or neutral emotions. It can also be applied to social platforms such as Twitter to help individuals or companies understand public sentiment. In the medical field, it is essential to comprehend patient sentiments to improve service quality and patient satisfaction.

However, users' tweets may contain many various aspects. For instance, "Suppose every single person in UK gets vaccinated with Pfizer's 90\% effective vaccine (no chance) - do you think UK govt will relax social distancing, family gathering, and mask wearing laws? We’ll still have to protect the 10\%," where the attitude towards 'Pfizer' is neutral, while the attitude towards the UK government is negative. The goal of aspect-based sentiment analysis, or ABSA, is to pinpoint the precise polarity toward a given feature \citep{pontiki2014aspect}. Generally, this task is formulated as predicting the polarity of the provided (sentence, aspect) pairs.

Given the inefficiency of manual feature refinement \citep{jiang2011target}, aspect-based sentiment classification primarily adopts neural network methods. Utilizing Convolutional Neural Networks (CNN), Recurrent Neural Networks (RNN), and Long Short-Term Memory networks (LSTM), researchers can more effectively capture complex features and dependencies within text data \citep{xing2019convolutional}.

In 2017, Vaswani et al. introduced the Transformer model \citep{vaswani2017attention}, an architecture based on self-attention mechanisms, which achieves higher efficiency and effectiveness in processing sequence data compared to previous RNN and CNN models. Due to its parallel processing capabilities and efficient learning of long-distance dependencies, the Transformer model has become the foundation for many subsequent NLP tasks.

In 2018, researchers from Google proposed the BERT (Bidirectional Encoder Representations from Transformers) model, further advancing ABSA and other NLP tasks \citep{pontiki2014aspect}. BERT, by pre-training a large-scale bidirectional Transformer encoder to learn text representations, achieved significant performance improvements across various NLP tasks, as demonstrated by Devlin [5], including sentiment analysis.

Pre-trained language models, especially BERT and its variants (such as RoBERTa, ALBERT, etc.), have provided new approaches to ABSA. An increasing number of researchers have begun to further explore BERT.

Despite the great success of BERT, its potential in specific application domains, such as ABSA in medical social media texts, remains underexplored. Since understanding patients' emotions in this field is crucial for improving service quality and patient satisfaction, this paper focuses on the medical domain and develops different deep learning architectures in combination with BERT and other pre-trained models to improve their performance in this challenging task. This will involve a detailed study of methods such as Fully Connected Networks (FCN), Convolutional Neural Networks (CNN) and Graph Convolutional Networks (GCN), each chosen for their unique advantages in capturing textual information.

The main contributions of this paper are as follows:

1) This paper designs model architectures that cater to a range of deep learning models, aligning with the process from feature extraction to sentiment classification.

2) The paper analyzes the varying effects of coupling pre-trained models with different neural network architectures, providing a research benchmark for sentiment analysis with limited samples in the medical field.

3) Experiments demonstrate that fine-tuning pre-trained models on specific texts significantly enhances performance.

The other portions of this work are organized as follows:

The study begins by outlining the experimental approach, which includes a brief description of the tasks, datasets, deep learning models, and training procedure. The experimental findings are then demonstrated, and the effects of various neural network topologies are examined, along with the causes of performance disparities, are compared. Finally, the conclusion section discusses the impact of the work on future research and the potential of the BERT model in understanding public sentiment in the medical domain.

\section{Method}
\label{sec:headings}
This experiment completes the task of aspect-level target sentiment classification in two steps. Firstly, aspects and corresponding sentiments are extracted from the dataset to allow the model to read and predict, according to the requirements of the task. Secondly, features are extracted from the textual content using two different pre-trained BERT models. Sentiment classification is performed using one of several classic deep learning models, including FCN, CNN, or GCN.

The following sections will provide detailed discussions of the experimental task, the dataset used in the experiment, the model structure, and the specific design architecture of the deep learning network.

\subsection{Task Description}
In the task of sentiment identification within Aspect-Based Sentiment Analysis (ABSA), a sentence $ S $ consists of $ N $  words: $ \{ s_1,s_2,s_3,\ldots s_N \} $ , and a set of predefined targets, also known as aspects, represented by the list $ A $ : $ \{ a_1,a_2,a_3,\ldots a_k \} $ , where the size of the aspect list is $ k $. Each aspect $ a_i=\{ s_{i_1},s_{i_2},s_{i_3} ,\ldots,s_{i_M} \} $ is a sub sequence of the sentence $ S $ and contains $ M $ words, where $ M\in\ [1,\ N) $.  Sentiment polarity of sentence $ S $ toward each feature $ a_i $, including positive, neutral, and negative attitudes, is assessed by aspect-level sentiment categorization. This goal corresponds with Task 4 of SemEval-2014 \citep{pontiki2014aspect}.

\begin{table}[h!]
\centering
	\caption{An instance of an aspect-based sentiment analysis assignment.}
\resizebox{\columnwidth}{!}{%
\begin{tabular}{@{}ll@{}}
\toprule
\multicolumn{1}{c}{\textbf{Tweet}}                                                  & \multicolumn{1}{c}{\textbf{Sentiments}} \\ \midrule
Today's best moment was watching a new patient in the waiting room read her         & COVID vaccine: Neutral                  \\
son his new book after she got her first COVID vaccine. Thank you @rorcarolinas     & rorcarolinas: Positive                 \\
for the books, and @DurhamHealthNC for the vaccine transfers !                      & DurhamHealthNC: Positive         \\ \bottomrule
\end{tabular}%
}
\label{111}
\end{table}

Table~\ref{111} shows an example of the task. The task is to input the sentences of the tweets into the model with the corresponding aspect location, and the model will output one of three sentiments: positive, negative, or neutral, after analysis and calculation.

\subsection{Dataset Description}

The experiment uses the METS-CoV dataset proposed by \citep{zhou2020metscov}. This dataset contains medical entities and their corresponding sentiments related to COVID-19 tweets. The dataset comprises a significant amount of text data collected from social media platforms such as Twitter. It is known for its high timeliness and public engagement, reflecting the public's attitudes and views on various topics during the COVID-19 pandemic.

\begin{figure}[h!]
	\centering
	\includegraphics[width=0.95\textwidth]{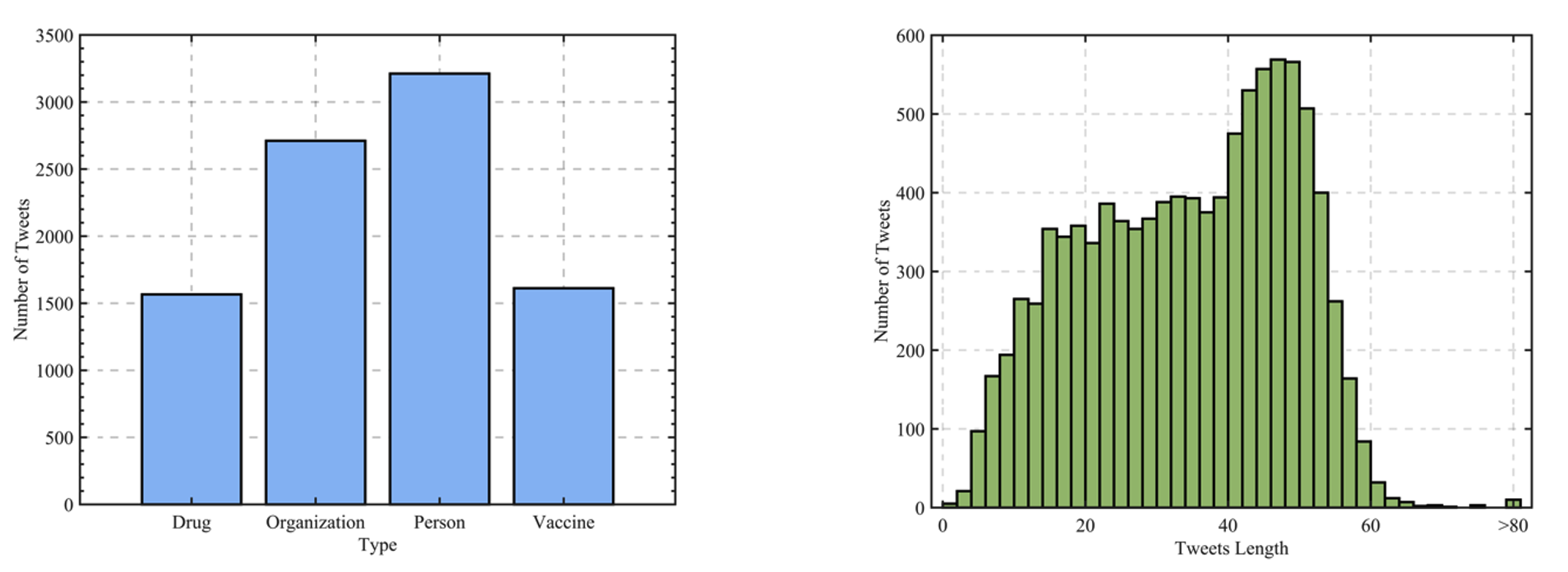} 
	\caption{The distribution of the types and tweets lengths in the dataset METS-CoV.}
	\label{fig:fig0}
\end{figure}

This dataset comprises approximately 10,000 tweets covering various aspects of the COVID-19 pandemic, including vaccination, drug treatment, and public health measures. The data was restructured based on the sentiments and aspects contained in the sentences, resulting in a total of 9,000 data points for model training. The data points consist of the original text, the corresponding sentiment attitude, and the specific aspect. The distribution of these aspects is illustrated in Figure \ref
{fig:fig0}.

From the distribution of data, it is clear that the dataset has a significant presence in the Person and Organization categories, allowing the model to thoroughly learn the public's views and attitudes towards different social entities during the pandemic. The dataset also includes a substantial amount of content related to drugs and vaccines, highlighting its relevance in the fields of public health and medicine. This text reflects public concern over health issues and the demand for medical information, as well as the response to medical policies during a global health crisis. The sentence lengths in the dataset mainly range from 10 to 50 words, which conforms to the common characteristics of social media texts: concise and direct. This text length is suitable for rapid information exchange, reflecting the speed and convenience of information dissemination on social media platforms.

Overall, this dataset illustrates the nuanced variations in public sentiment during the pandemic and emphasizes the significance of public health topics in social discourse. It is appropriate for conducting a thorough analysis of public emotions on social media in the context of the pandemic, particularly when discussing people's responses to social events, public figures, medical policies, and health information. This dataset is an excellent resource for analyzing and training BERT-based sentiment classification models, especially for recognizing and interpreting public emotions on social media. It demonstrates the performance differences in handling real-time social media data when combining the BERT model with other deep learning models.

\subsection{Model Design}

\begin{figure}[h!]
	\centering
	\includegraphics[width=0.45\textwidth]{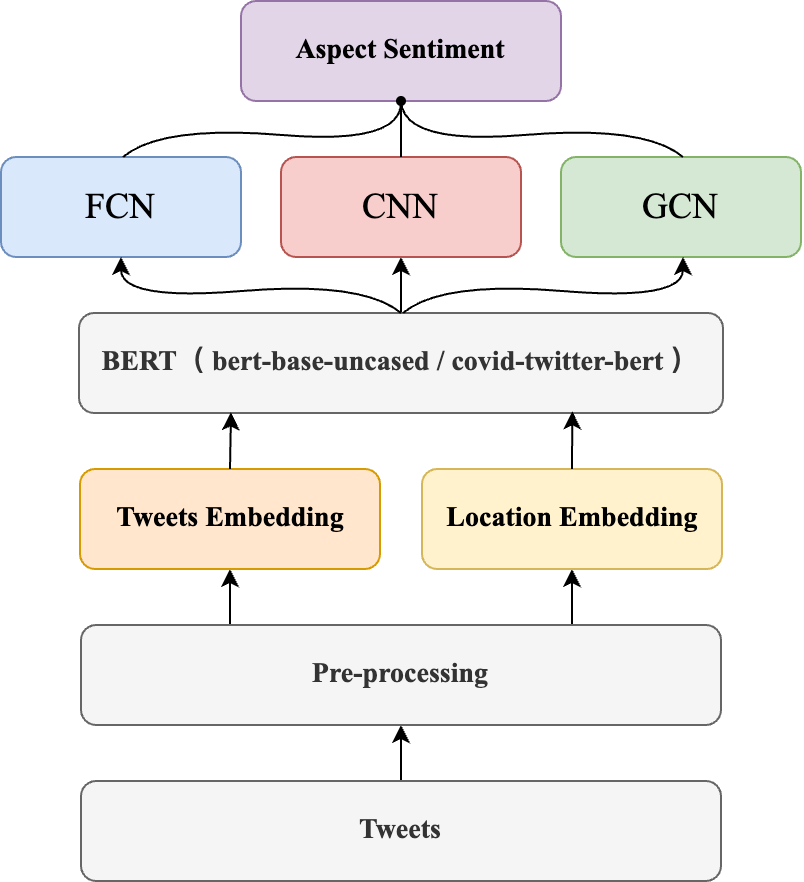} 
	\caption{The architecture of the model.}
	\label{fig:fig1}
\end{figure}

The model architecture used in this experiment consists of three parts: the input layer, the BERT layer, and the deep learning layer. The detailed procedure is depicted in Figure \ref{fig:fig1}.

The input layer processes the original data after it has been cleaned. Depending on the task's requirements, it segments the text to extract the content to the left and right of the target (aspect) and the target (aspect) itself. And then the tokenizer of the BERT model is used to convert the text into sequences, creating two embedding inputs to suit the ABSA task.

To better adapt the general model to the dataset used in the experiment, which contains a large amount of text from Twitter centered around COVID-19, the BERT layer in the model is replaced by COVID-TWITTER-BERT model and trained again. Contrasting with the original BERT-large-uncased model, COVID-TWITTER-BERT has been pre-trained on a corpus of messages about COVID-19 on Twitter, enabling it to demonstrate enhanced downstream performance in this experiment and provide more results for us to analyze \citep{muller2023covid}.

In the deep learning layer, to better compare the performance of different deep learning models in sentiment classification using the features extracted by the BERT model, this experiment has designed three deep learning models. These are the fully connected layer, convolutional neural network, and graph neural network. The specific designs of these models will be detailed in subsequent sections.

\subsection{Design of Deep Learning Models}

To ensure accurate determination of aspect-level sentiments and evaluate the natural language processing capabilities of different deep learning models on small texts. This experiment employs three types of deep learning models: fully connected networks, convolutional neural networks, and graph convolutional neural networks. The following sections introduce the structural features of these networks and the design rationale behind their implementation in this experiment.

  Fully connected network (FCN): FCN is a basic neural network structure where each neuron is connected to all neurons in the previous layer \citep{huang2018densely}. For this experiment, a two-layer fully connected network was applied. This is because a single layer has limited data processing capability, while too many layers increase computational complexity and risk gradient vanishing or exploding. The network comprises of two layers, $D_1\in\mathbb{R}^{300\times D}$ and $ D_2\in\mathbb{R}^{3\times300}$, where $ D $ is determined by the dimension of the BERT model, and all of them are connected through the ReLU activation function. This simple and efficient structure can capture non-linear relationships in data. In aspect-level sentiment analysis, the fully connected network aids in understanding the overall sentiment in the text but may not capture fine-grained information.
  
Convolutional Neural Network (CNN): CNN, commonly used in image processing, is also widely applied in text analysis  \citep{albawi2017understanding}. The text describes the use of convolutional layers to extract local features in text data, which is useful for capturing key information in text segments. The experiment employs two one-dimensional convolutional layers to identify and extract key expressions and sentiment words related to specific aspects. The parameter count is similar to other models. CNNs are effective at processing local information in text, but they may have a weaker understanding of complex semantic relationships.

Graph Convolutional Network (GCN): GCN is a deep learning model that specifically designed to process graph-structured data \citep{zhang2019graph}. In natural language processing, GCN can capture dependencies between words, which is crucial for understanding complex semantic structures. In this experiment, text is treated as a graph structure, where nodes represent words and edges represent relationships between words. Two layers of GCN, each sized $ D^2$, are used. At the same time, based on the dataset, this experiment generates relationships that enable the model to capture the deeper semantics in the text and thus understand more accurately the different aspects of emotional tendencies.

In summary, each of these models has its own advantages and characteristics. The fully connected network has a simple structure, making it easy to implement and train. CNN, on the other hand, is excellent at processing local features and is suitable for capturing key information in text segments. Meanwhile, GCN is effective at handling complex semantic relationships in text.

After being designed and integrated with BERT model, all three models have a parameter training volume of around 100M, resulting in minimal parameter differences between them. By comparing the performance of these three models on a small text dataset, I can more comprehensively assess their effectiveness in aspect-level sentiment analysis tasks.

\section{Experiment}
\subsection{Model Training}
In this experiment, the author employed the bert-base-uncased model from the BERT pre-training suite and a specially enhanced COVID-TWITTER-BERT for fine-tuning. The number of Transformer blocks is set to 12, with the size of BERT hidden layers being 768 for the bert-base-uncased model and 1024 for COVID-TWITTER-BERT, and the number of attention heads is 12. The total number of parameters in the pre-trained models is around 100M. During fine-tuning, the author maintained a dropout probability of 0.1, an epoch count of 20, an initial learning rate of 2e-5, and a batch size of 16. The dataset was divided into training, validation, and test sets at a ratio of 70:30:30. Every epoch of the training procedure concluded with an evaluation of the validation set's performance. In order to prevent model overfitting, an early stopping mechanism was implemented, whereby training was stopped after 5 epochs if performance declined.

\subsection{Results}

\subsubsection{Performance Metrics}

This experiment employed several common performance evaluation metrics, such as Accuracy and F1 Score. These metrics help to comprehensively assess the model's performance in aspect-level sentiment classification tasks. The table displays the performance of each model on this dataset, with data results calculated from 5 random experiments, i.e., mean ± standard deviation. The mean is calculated by averaging the performance under different seeds. The standard error (std) is calculated by dividing the standard deviation of the mean by the square root of the number of seeds.

\subsubsection{Model Comparison}
Table \ref{table2} illustrates the performance disparities of various models used in this experiment on the dataset. The following data can be observed.

\begin{table}[h!]
\centering
\caption{The results of the experiments.}
\setlength{\tabcolsep}{12pt} 
\begin{tabular}{@{}cccc@{}}
\toprule
\textbf{Model} & \textbf{Pretrained Model} & \textbf{ACC}          & \textbf{F1}           \\ \midrule
FCN            & bert-base-uncased         & 65.99 ± 0.77          & \textbf{53.52 ± 1.70} \\
CNN            & bert-base-uncased         & \textbf{66.29 ± 0.64} & 48.46 ± 2.60          \\
GCN            & bert-base-uncased         & 63.77 ± 0.43          & 45.26 ± 0.81          \\ 
FCN            & COVID-TWITTER-BERT        & 66.79 ± 1.20          & 46.70 ± 2.72          \\
CNN            & COVID-TWITTER-BERT        & \textbf{72.97 ± 1.32} & \textbf{65.37 ± 1.54} \\
GCN            & COVID-TWITTER-BERT        & 70.73 ± 1.07          & 60.72 ± 2.39          \\ \bottomrule
\end{tabular}
\label{table2}
\end{table}

FCN: The results indicate that FCN performs well in processing overall sentiment tendencies, but is somewhat inadequate in capturing and distinguishing fine-grained sentiment information.

CNN: CNN excels in extracting local textual features, especially in identifying key sentiment vocabulary.

GCN: GCN exhibits weaker performance in understanding complex relationships between vocabularies and is less effective in handling the output content of pre-trained models.

Moreover, substituting the original bert-base-uncased base model with the COVID-TWITTER-BERT model, which has been modified for Twitter-related texts, has resulted in a certain improvement in the performance of all models.

\subsubsection{Analysis of Results}
In this experiment, the authors conducted a concise analysis of three distinct neural networks: Graph Convolutional Networks (GCN), Fully Connected Networks (FCN), and Convolutional Neural Networks (CNN). The specific findings are as follows:
\begin{itemize}
	\item Fully Connected Networks (FCN)
 
 FCNs are known for their simplicity and efficiency, demonstrating stable performance across various tasks. With the bert-base-uncased pre-trained model, FCNs, which serve as versatile universal function approximators, can learn non-linear patterns in the data and outperform GCNs in terms of accuracy and F1 scores, demonstrating better overall sentiment understanding. However, when switching to the pre-trained COVID-TWITTER-BERT model, the F1 scores of the FCNs decrease significantly. Analysis of the model's predictions reveals a reduced ability to discriminate between positive and negative sentiments, suggesting that FCNs, due to their simple and intuitive structure, may lack the intrinsic ability to capture certain complex data features, such as the logical and sequential features in textual data. The structural simplicity may limit the model's ability to learn task-relevant abstract and latent patterns when transitioning from models such as COVID-TWITTER-BERT, thereby affecting performance on specific tasks, especially sentiment analysis tasks that require fine-grained understanding.

	\item Convolutional Neural Networks (CNN)

 CNNs are widely favored in text processing tasks because of their exceptional ability to extract local features. Experimental results show that CNNs outperform all other networks, regardless of whether the bert-base-uncased or the COVID-TWITTER-BERT pre-trained model is used. In particular, when combined with the Twitter pre-trained model, CNNs not only achieve the highest accuracy, but also see a significant increase in F1 scores, demonstrating their powerful ability to capture emotional tendencies and local semantic information in social media texts. An advantage of the CNN architecture over other architectures is its ability to automatically learn useful contextual structure information. By using stacks of deep convolutional layers, CNNs can capture different sizes of contextual patterns and the intrinsic hierarchical structure of such features. In addition, compared to FCNs, CNNs are highly efficient in their use of parameters. This means that CNNs can learn more information with the same number of parameters.
 
	\item Graph Convolutional Networks (GCN)

 GCNs have demonstrated exceptional performance in processing data characterized by complex relationships and structures, particularly in capturing dependencies among words. However, when utilizing the bert-base-uncased pre-trained model, the performance of GCNs appears relatively weaker. A significant enhancement in GCN's performance is observed upon switching to a pre-trained model optimized for Twitter. Research suggests that this improvement may be attributed to the pre-trained model's ability to automatically capture dependencies between words. These captured dependencies, however, may not align perfectly with manually designed dependency rules, leading to minimal or even negative impacts when further refining with a GCN model based on the original text. This indicates that Graph Convolutional Neural Networks are less suited for processing outputs extracted by large models. Instead, positioning them at the input stage to apply the original text's dependency relationships more effectively aids in machine learning.
 
\end{itemize}

In summary, the CNN architecture demonstrates a better capability to learn from the output of pre-trained models. While FCNs exhibit good stability, they require improvements. GCNs show the weakest performance among the evaluated networks.

\section{Conclusion}
\label{sec:others}
In this paper, the author conduct extensive experiments to investigate methods for sentiment classification tasks in medical-related texts, as well as the performance enhancement of combining different deep learning models with pretrained models. The experiments reveal that: 

1)	The role of architecture: Three different deep learning models, when combined with pretrained models, exhibit varying performances. CNNs perform exceptionally well in sentiment analysis tasks of social media texts, especially when used in conjunction with pretrained models optimized for social media texts. While CNNs show robust performance with general-purpose pretrained models, they require further optimization in specific domains. The performance of GCNs mainly depends on the adopted pretrained models and struggles to significantly leverage the output of these models.

2)	The enhancement effect of BERT: The results indicate that further training BERT on domain-specific corpora can significantly improve model performance in text classification tasks. 

On a dataset like METS-CoV, which is relatively small and revolves around medical themes, the combination of pretrained models with a CNN architecture demonstrates strong performance. These findings provide valuable insights for future selection of suitable neural network and pretrained model combinations in specific tasks.

The discoveries not only offer new perspectives for research in sentiment analysis within the medical field but also pave new paths for future NLP applications and research. I hope to facilitate more participation in the research of natural language processing of medical texts, and I aim to delve deeper into the workings of BERT for greater breakthroughs.

\bibliographystyle{unsrtnat}
\bibliography{references}  






\end{document}